\documentclass[letterpaper]{article} % DO NOT CHANGE THIS
\usepackage{aaai25}  % DO NOT CHANGE THIS
\usepackage{times}  % DO NOT CHANGE THIS
\usepackage{helvet}  % DO NOT CHANGE THIS
\usepackage{courier}  % DO NOT CHANGE THIS
\usepackage[hyphens]{url}  % DO NOT CHANGE THIS
\usepackage{graphicx} % DO NOT CHANGE THIS
\usepackage{natbib}  % DO NOT CHANGE THIS AND DO NOT ADD ANY OPTIONS TO IT
\usepackage{caption} % DO NOT CHANGE THIS AND DO NOT ADD ANY OPTIONS TO IT
\usepackage{algorithm}
\usepackage{algorithmic}

\usepackage{newfloat}
\usepackage{listings}
\DeclareCaptionStyle{ruled}{labelfont=normalfont,labelsep=colon,strut=off} % DO NOT CHANGE THIS
\floatstyle{ruled}
\newfloat{listing}{tb}{lst}{}
\floatname{listing}{Listing}
\usepackage{bm}
\usepackage{amsmath, amssymb}
\usepackage{multirow}
\usepackage{makecell}
\usepackage{graphicx}
\usepackage{booktabs}
\usepackage{subfigure}
\newtheorem{definition}{Definition}
\newtheorem{theorem}{Theorem}

\title{Does GCL Need a Large Number of Negative Samples? Enhancing Graph Contrastive Learning with Effective and Efficient Negative Sampling}
\author{
    Yongqi Huang\textsuperscript{\rm 1}\equalcontrib,
    Jitao Zhao\textsuperscript{\rm 1}\equalcontrib,
    Dongxiao He\textsuperscript{\rm 1}\thanks{Corresponding Author.},
    Di Jin\textsuperscript{\rm 1},
    Yuxiao Huang\textsuperscript{\rm 2},
    Zhen Wang\textsuperscript{\rm 3}
}
\affiliations{
    \textsuperscript{\rm 1}College of Intelligence and Computing, Tianjin University, Tianjin, China\\
    \textsuperscript{\rm 2}Department of Data Science, George Washington University, Washington, D.C., U.S.A.\\
    \textsuperscript{\rm 3}School of Cybersecurity, Northwestern Polytechnical University, Xi’an, China\\
    \{yqhuang, zjtao, hedongxiao, jindi\}@tju.edu.cn, yuxiaohuang@gwu.edu, w-zhen@nwpu.edu.cn
}

\begin{document}

\maketitle

\begin{abstract}
Graph Contrastive Learning (GCL) aims to self-supervised learn low-dimensional graph representations, primarily through instance discrimination, which involves manually mining positive and negative pairs from graphs, increasing the similarity of positive pairs while decreasing negative pairs. 
Drawing from the success of Contrastive Learning (CL) in other domains, a consensus has been reached that the effectiveness of GCLs depends on a large number of negative pairs. As a result, despite the significant computational overhead, GCLs typically leverage as many negative node pairs as possible to improve model performance.
However, given that nodes within a graph are interconnected, we argue that nodes cannot be treated as independent instances. Therefore, we challenge this consensus:  \textit{Does employing more negative nodes lead to a more effective GCL model?}
To answer this, we explore the role of negative nodes in the commonly used InfoNCE loss for GCL and observe that: (1) Counterintuitively, a large number of negative nodes can actually hinder the model's ability to distinguish nodes with different semantics. (2) A smaller number of high-quality and non-topologically coupled negative nodes are sufficient to enhance the discriminability of representations.
Based on these findings, we propose a new method called GCL with Effective and Efficient Negative samples, E2Neg, which learns discriminative representations using only a very small set of representative negative samples. E2Neg significantly reduces computational overhead and speeds up model training. 
We demonstrate the effectiveness and efficiency of E2Neg across multiple datasets compared to other GCL methods.

\end{abstract}

\begin{links}
    \link{Code}{https://github.com/hedongxiao-tju/E2Neg}
\end{links}

\section{Introduction}
Graph Neural Networks (GNNs) \cite{GCN, GraphSAGE} have emerged as a powerful strategy for graph mining, aiming to encode structured high-dimensional graphs into low-dimensional embeddings. GNNs are widely applied in various scenarios such as social recommendation \cite{SocialRecommendation}, community detection \cite{CommunityDetectionSurvey, JinW21}, and sarcasm detection \cite{Wang23}.
% anomaly detection \cite{AnomalyDetection}.
Training high-quality GNNs typically requires an extensive amount of labels, which is labor-intensive and costly. 
% Considering the inherent scarcity of labels, the label dependency problem significantly limits the applicability of GNNs.
The scarcity of labels limits the applicability of GNNs due to label dependency.

% 介绍GCL
Graph Contrastive Learning (GCL) has recently gained significant attention for its ability to self-supervisedly train GNNs, producing expressive representations. Inspired by the success of contrastive learning in other domains, most GCLs are based on instance discrimination \cite{GRACE, GCA}. Specifically, they first mine positive and negative samples, then train the model using the InfoNCE loss \cite{InfoNCE}, which maximizes the similarity between positive pairs while minimizing the similarity between negative pairs. For an anchor node, positive samples are typically constructed through augmentation to align the invariant semantics of the perturbed data, while negative samples are often chosen from other nodes to enhance the discriminability of the representations.

% 引出采样工作
The effectiveness of instance-discrimination-based GCLs highly depends on the sampling strategy. ProGCL \cite{ProGCL} estimates the probability of negatives being true negatives by using a Beta Mixture Model (BMM), and can further leverage this probability to weight the negatives and synthesize new negatives. Local-GCL \cite{LocalGCL} removes first-order neighbors from negatives to positives based on graph homophily assumption. HomoGCL \cite{HomoGCL} estimates the probability of neighbors being true positives using graph homophily and calculates node similarity through soft clustering, using it as a weight for positives. $\text{B}^{2}$-Sampling \cite{B2-sampling} selects representative negatives by designing a balanced sampling strategy and corrects the labels which can be hard to learn. 
% 引出共识
Although these methods improved the sampling strategy and achieved some success, all these methods share a consensus that incorporating more negative samples improves model performance. This consensus has been discovered and validated by many researchers in Computer Visions (CV) \cite{MoCo, SimCLR} and other domains.
%In the field of Computer Vision (CV), certain studies such as MoCo \cite{MoCo} and SimCLR \cite{SimCLR}) have shown that enhancing the number of negatives improves the effectiveness of model. Previous studies on graph have not provided evidence or validation for this claim. They all make a same common assumption when defining samples: all pairs of samples, except those from the same nodes in different views, are regarded as negatives.

% These methods share a consensus that incorporating more negative samples improves model performance. This consensus has been discovered and validated by many researchers in computer visions \cite{MoCo, SimCLR}.

However, a gap has emerged regarding this consensus between graph and other domains: This consensus is based on the premise that instances are independent of each other. This premise is valid within the CV. In graph data, the presence of topological connections between nodes results in instances that do not satisfy the independence premise. 
This raises an important question: \textit{\textbf{Does employing more negative nodes lead to a more effective GCL model?}} 

To answer this question, 
we theoretically investigate the role of topologically coupled negative nodes in the InfoNCE loss. We first demonstrate that within the topological receptive field of graph encoders, multiple topologically coupled node representations share the same semantics with only minor differences, thereby violating the assumption of instance independence. Further analysis reveals some interesting and counter-intuitive findings: 1) A large number of negative sample nodes can actually hinder the model’s ability to learn semantic distinctions between nodes. 2) For node-level tasks, a small number of representative negative samples are sufficient to produce distinguishable representations.

Based on these findings, we propose a new GCL method called GCL with Effective and Efficient Negative samples (E2Neg), which enables the GNN encoders to learn discriminative representations using only a very small number of representative negative samples (typically fewer than 50). Specifically, E2Neg preprocesses the graph using spectral clustering, selects representative nodes based on centrality, and reconstructs subgraphs around these nodes to identify topologically decoupled and representative samples. We further design a data augmentation for these negative samples, aligning the semantics of nodes that are topologically coupled with them. Additionally, by training with only a small selection of samples, E2Neg significantly reduces computational overhead and boosts training speed.
%Specifically, we preprocess the graph through spectral clustering, then select representative nodes based on centrality, and reconstruct subgraphs for these nodes. By using only representative nodes for training the objective function, the model can effectively distinguish nodes with different semantic information. Our method reduces the computational overhead and significantly accelerates model training.

Our contributions can be summarized as follows:
\begin{itemize}
    \item
% we explore the role of negative samples in InfoNCE，and find 大量的负例节点不会提升模型的性能，反而会影响模型区分不同语义节点的判别能力。以及仅仅使用一小部分节点作为负样本参与训练are sufficient to enhance the discriminability of representations.
We explore the function of negative samples in InfoNCE and find that increasing the number of negatives does not improve the performance of the model. Instead, it may hinder the model's ability to distinguish nodes with different semantics. Moreover, using only a very small subset of nodes as negatives is sufficient to enhance the discriminability of representations.
    \item 
% 该模型仅通过一个非常小的代表性的负样本集训练模型。E2Neg减减轻了计算开销，并且能够显著加速模型的训练。
Based on these findings, we propose a novel GCL method called GCL with Effective and Efficient Negative samples (E2Neg), where training only requires a very small set of representative negatives. E2Neg reduces computational overhead and significantly accelerates model training.
    \item 
We conduct extensive experiments on multiple datasets. Experimental results demonstrate the effectiveness and efficiency of E2Neg.
\end{itemize}
%77:Preliminaries
\section{Preliminaries}
\subsubsection{Notations.}
Given a graph $\mathcal{G}=(\mathcal{V},\mathcal{E})$, where $\mathcal{V}=\{ v_1, v_2, \cdots, v_N \}$ is the set of nodes, $|\mathcal{V}|=N$, and $\mathcal{E} \subseteq \mathcal{V} \times \mathcal{V}$ is the set of edges. $\bm{X} \in \mathbb{R}^{N \times F}$ is the feature matrix, $\bm{X}_i \in \mathbb{R}^{F}$ is the feature of $v_i$. $\bm{A} \in \{0, 1\}^{N \times N}$ is the adjacency matrix, and $\bm{A}_{ij}=1$ iff $(v_i, v_j) \in \mathcal{E}$. The normalized Laplacian matrix is defined as $\bm{L}=\bm{I}_{N}-\bm{D}^{-\frac{1}{2}}\bm{A}\bm{D}^{-\frac{1}{2}}$, where $\bm{I}_{N}$ is an identity matrix and $\bm{D}$ is diagonal degree matrix with $\bm{D}_{ii} = \sum_{j=1}^{N}\bm{A}_{ij}$ for $i \in \mathcal{V}$. 

\subsubsection{Eigenvector and Eigenvalue.} The normalized graph Laplacian can be decomposed as $\bm{L}=\bm{U}\Lambda\bm{U}^{\top}$, where $\bm{U}=[\bm{u}_{1}, \bm{u}_{2}, \cdots ,\bm{u}_{N}] \in \mathbb{R}^{N\times N}$ is the eigenbasis, which consists of a set of eigenvectors. $\Lambda=diag(\{ {\lambda_{i}}\}_{i=1}^{N})$ are the eigenvalues. As for each eigenvector $\bm{u}_{i} \in \mathbb{R}^{N}$ has a corresponding eigenvalue $\lambda_{i}$. Without loss of generality, assume $0 \leq \lambda_{1} \cdots \leq \lambda_{N} \leq 2$. 

\subsubsection{Graph Contrastive Learning.} Our objective is to train an encoder $f$ to get representation $\bm{H}=f(\bm{A}, \bm{X})\in\mathbb{R}^{N\times D}$ without label information. Augmented graphs are typically generated through augmentation techniques, such as edge dropping and feature masking. These augmented graphs are fed into $f$, which generates the representations $\bm{H}^{\alpha}$ and $\bm{H}^{\beta}$ for the two views. The objective function is defined by InfoNCE \cite{InfoNCE} loss as:
\begin{equation}
\begin{aligned}&
\ell\left(\bm{h}^{\alpha}_i,\boldsymbol{h}^{\beta}_i\right)=\\&\log\frac{e^{\theta(\boldsymbol{h}^{\alpha}_i,\boldsymbol{h}^{\beta}_i)/\tau}}{e^{\theta(\boldsymbol{h}^{\alpha}_i,\boldsymbol{h}^{\beta}_i)/\tau}+\sum_{k\neq i}e^{\theta(\boldsymbol{h}^{\alpha}_i,\boldsymbol{h}^{\beta}_k)/\tau}+\sum_{k\neq i}e^{\theta(\boldsymbol{h}^{\alpha}_i,\boldsymbol{h}^{\alpha}_k)/\tau}},
\end{aligned}
\end{equation}
where $\theta(\cdot)$ denotes the cosine similarity function, with $\theta(\boldsymbol{h}^{\alpha}_i, \boldsymbol{h}^{\beta}_i)$ representing the similarity between the same node in different views, which constitutes positive samples. Conversely, $\theta(\boldsymbol{h}^{\alpha}_i, \boldsymbol{h}^{\beta}_k)$ represents the similarity between different nodes in different views, and $\theta(\boldsymbol{h}^{\alpha}_i, \boldsymbol{h}^{\alpha}_k)$ represents the similarity between different nodes in the same view, both of these cases correspond to negative samples.

%77:Analysis of Sampling in InfoNCE
\section{Negative Sampling Analysis}
% 在这部分，我们从理论上分析了在InfoNCE损失中的负样本，我们对其工作方式在损失函数上进行了分析，发现了现有定义的负样本对，随着训练会减弱样本的区分度。接着，我们根据该缺陷，提出了对应的采样方案得以解决。
In this section, we theoretically analyze negative samples in the InfoNCE loss. We observe that the current definition of negatives tends to diminish the distinctiveness of samples during train. To address this issue, we propose a new sampling strategy that effectively mitigates this limitation.

\begin{definition}[Receptive Field $\mathcal{R}$]
    The receptive field of $v_{i}$ is defined as the set of nodes whose aggregated information influences the representation of node $v_{i}$ during the propagation process. It can be formulated as:
    \begin{equation}
        \mathcal{R}(v_{i}) = \bigcup_{t=0}^{\hat{k}} \{ v_{j} \in \mathcal{V} \mid \text{dist}(v_{i}, v_{j}) = t \text{ and } w_{t}(v_{i}, v_{j}) > \xi \},
    \end{equation}
    where $\hat{k}$ is the number of propagation layers, $\text{dist}(v_{i}, v_{j})$ represents the shortest path distance between nodes $v_{i}$ and $v_{j}$, $w_{t}(v_{i}, v_{j})$ is the weight of the information aggregated from $v_{j}$ to $v_{i}$ at layer $t$, and $\xi$ is a threshold for significant contributions. The receptive field expands as the number of layers increases, incorporating a broader range of nodes that contribute to the representation of $v_{i}$.
\end{definition}
Specifically, if the number of propagation layers is $\hat{k}$, the receptive field of $v_{i}$ includes all nodes within a distance of $\hat{k}$ that significantly contribute to $\bm{h}_{i}$. $w_{t}(v_{i}, v_{j})$ quantify this contribution, with higher values indicating a greater impact on the representation $\bm{h}_{i}$.

\begin{definition}[Semantic Block $\mathcal{S}$]
    A semantic block $\mathcal{S}$ is defined as a set of nodes within a graph that form a cohesive group centered around a core semantic \(s_c\). This block $\mathcal{S}$ can be formulated as:
    \begin{equation}
        \mathcal{S} = \{ v_{j} \in \mathcal{V} \mid \| x_j - s_c \| \leq \delta \},
    \end{equation}
    where $s_c$ represents the core semantic of $\mathcal{S}$, and $\delta$ is a small value determining the allowable difference between the nodes and $s_c$.
\end{definition}
A corresponding $\mathcal{S}$ exists for each graph node. These semantic blocks do not overlap, meaning that different semantic blocks are non-coupled.
% 对于一张图中的每一个节点，其均存在一个对应的语义块\mathcal{S}。并且多个语义块之间是不存在交叉的，即不同语义块间是非耦合的。

\begin{theorem}[Semantic Block-Based Decomposition]
    Assume that a graph contains $k$ semantic blocks $\mathcal{S}=\{\mathcal{S}_{1}, \cdots, \mathcal{S}_{k} \}$, corresponding to a set of core semantics $s = \{s_1, \dots, s_k\}$, where each $s_j$ is the core semantic associated with the block $\mathcal{S}_j$. If an anchor node $v_i$ belongs to the semantic block $\mathcal{S}_j$, $x_i$ can be decomposed as:
    \begin{equation}
        x_i = s_j + \epsilon_i ,
    \end{equation}
    where $s_j$ is the core semantic of the block $\mathcal{S}_j$, and $\epsilon_i$ represents the individual deviation of $v_i$ from this core semantic.
\end{theorem}

The feature difference between nodes varies depending on whether they belong to the same semantic block. For nodes within the same semantic block, the difference is determined by the individual deviations from the core semantic. For nodes belonging to different semantic blocks, the difference is influenced by the core semantic associated with each block. Given two different semantic blocks $\mathcal{S}_p$ and $\mathcal{S}_q$, it can be expressed as:
\begin{equation}
\begin{aligned}
\bigtriangleup \text{diff}_{\text{intra}} &= \| \epsilon_i - \epsilon_j \|, && \text{if } v_i, v_j \in \mathcal{S}_p , \\
\bigtriangleup \text{diff}_{\text{inter}} &= \| s_p - s_q \|, && \text{if } v_i \in \mathcal{S}_p \text{ and } v_j \in \mathcal{S}_q ,
\end{aligned}
\end{equation}
where $ \bigtriangleup\text{diff}_{\text{intra}} $ represents the difference between negative samples within $\mathcal{S}_p$. An increase in this difference corresponds to a decrease in the similarity of negative samples within the same semantic block. Conversely, $ \bigtriangleup\text{diff}_{\text{inter}} $ represents the difference between negative samples across $\mathcal{S}_p$ and $\mathcal{S}_q$. An increase in this difference is equivalent to reducing the similarity of negative samples between $\mathcal{S}_p$ and $\mathcal{S}_q$, precisely the behavior we desire from the model. This approach ensures that while distinguishing between semantic blocks, the model maintains a high similarity between node pairs within the same semantic block.
% where diff_intra代表了同一语义块内的负样本对的差异，该差异上升，其等价于降低同一个语义块内的负样本对的相似度，而diff_inter代表了不同语义块间的负样本对的差异，该差异上升等价于对应的负样本对的相似度降低。而后者正是我们希望模型能够做到的，并且在区分不同语义块的同时，保证相同语义块中的节点对，保持高的相似度。

\begin{theorem}[Sample Threshold Gradient Boundary]
Assume that during the optimization process, we consider two types of negative samples: inter- and intra-semantic block negatives, denoted as $\mathbf{N}_{\text{inter}}$ and $\mathbf{N}_{\text{intra}}$, respectively. During optimization, the gradients associated with these negatives are accumulated and compared. For the average similarity $\hat{\theta}(h_i,h_{j'})$ between samples from different semantic blocks, $P$ represents the number of sample pairs in the same semantic block, $\tau$ is the temperature coefficient. When 
$Pe^{1/\tau} = \sum_{j'=1,j'\neq i}^{N-P} e^{\theta(h_i,h_{j'})/\tau}$
, the model achieves a balance in distinguishing between $\mathbf{N}_{\text{inter}}$ and $\mathbf{N}_{\text{intra}}$.
\end{theorem}
% 当模型在区分$\mathbf{N}_{\text{inter}}$和$\mathbf{N}_{\text{intra}}$达到平衡时，存在一个比例系数 $\gamma = 1-\tau \ln^{(N-P)/P}$, where P为$\mathbf{N}_{\text{intra}}$的个数，\tau为温度系数
The model's discriminative focus is determined by the ratio of these gradients, which can be expressed as:
\begin{equation}
    \frac{\sum_{(u,v) \in \mathbf{N}_{\text{inter}}} \nabla \mathcal{L}(u,v)}{\sum_{(u,v) \in \mathbf{N}_{\text{intra}}} \nabla \mathcal{L}(u,v)} \geq \gamma,
\end{equation}
where $\gamma$ act as a boundary to determine whether the model focuses more on inter- or intra-semantic block distinctions. Define $\Delta \text{diff}_{\text{inter}} = \| s_p - s_q \|$ representing the semantic difference between inter-semantic block pairs and $\Delta \text{diff}_{\text{intra}} = \| \epsilon_u - \epsilon_v \|$ representing the variation within the same semantic block. Their differences quantify the distinctions within respective semantic blocks.
When $\Delta \text{diff}_{\text{intra}}$ increases, the model reduces the similarity within the same semantic block, encouraging node separation. When $\Delta \text{diff}_{\text{inter}}$ increases, the model decreases the similarity between different blocks, improving distinction across semantic blocks. $\tau$ helps balance these effects for effective block distinction while maintaining similarity within the same block.

We analyze the issues with existing sampling methods from the perspective of semantic blocks. To do so, we take the derivative of the InfoNCE loss for the similarity of negative samples, which yields the following expressions:
\begin{equation}
\frac{\partial\mathcal{L}_{\text{InfoNCE}}}{\partial \theta(h_i,h_{j})} = 
\frac{1}{\tau}
\frac{\phi(i,j)}{\phi(i, i')+
\sum_{k=1,k\neq i}^{n}
(\phi(i, k)+\phi(i, k'))},
\label{eqn:Derivation_1}
\end{equation}
where $\phi(i, j)=e^{\theta(h_i,h_j)/\tau}$, and $\theta(h_i,h_j)$ represents the gradient of any negative sample pair in the model, where $i$ and $j$ may belong to the same semantic block or different semantic blocks. 
% Summing the gradients of all negative samples for an anchor node $v_i$ allows us to decompose the total into loss contributions from pairs within the same semantic block and pairs across different semantic blocks, as follows:
By summing the gradients of all negative samples for an anchor node $v_i$, we can break down the total loss into contributions from pairs within the same semantic block and pairs across different semantic blocks, as follows:
% \theta(h_i,h_j)代表了模型中任意一对负样本的梯度，i和j可能属于同一个语义块，也可能不是。对锚节点 $v_i$ 的所有负样本的梯度求和，使我们能够将总梯度分解为来自同一语义块内对和不同语义块间对的损失贡献，具体如下：
\begin{equation}
\begin{aligned}
    \text{SG}(i)_{\text{intra}} &=\sum_{j=1,j\neq i}^{P} \frac{\partial\mathcal{L}_{\text{InfoNCE}}}{\partial \theta(h_i,h_{j})}, \\ \text{SG}(i)_{\text{inter}} &=\sum_{j'=1,j'\neq i}^{N-P} \frac{\partial\mathcal{L}_{\text{InfoNCE}}}{\partial \theta(h_i,h_{j'})},
\end{aligned}
\label{eqn:Derivation_2}
\end{equation}
where $\text{SG}(i)_{\text{intra}}$ represents the sum of the gradients for negatives within the same semantic block, $\text{SG}(i)_{\text{inter}}$ represents the gradients for negatives across different semantic blocks. $P$ represents the number of sample pairs in the same semantic block as $v_i$.
When the sum of gradients for negatives within the same semantic block exceeds that for negatives across different semantic blocks, the model tends to differentiate negatives within the same semantic block. Conversely, if the sum of gradients across different semantic blocks is greater, the model will focus on distinguishing negatives between different semantic blocks. To explore how the model switches between these targeting modes, we first assume that $\text{SG}(i)_{\text{intra}} = \text{SG}(i)_{\text{inter}}$. 
Through Eq. (\ref{eqn:Derivation_1}) and (\ref{eqn:Derivation_2}), we obtain the following expression: 
% 当同一个语义块中负样本的梯度和大于不同语义块中负样本的梯度和时，模型相对于要区分相同语义块内的负样本对，反之，模型会区分不同语义块的负样本对。为了探究模型对于这种目标模式的切换，这时我们先假设两部分是相等的，具体如下：
\begin{equation}
\begin{aligned}
    % \text{SG}(i)_{\text{intra}} &= \text{SG}(i)_{\text{inter}} \\
    \sum_{j=1,j\neq i}^{P} e^{\theta(h_i,h_j)/\tau} &= \sum_{j'=1,j'\neq i}^{N-P} e^{\theta(h_i,h_{j'})/\tau},    
\end{aligned} 
\end{equation}
% 由于在同一个语义块内，根据公式（5）样本间的差异为diff_intra，该值远小于diff_inter，故在这里我们假设\theta(h_i,h_j)=1，并设置\hat{\theta}(h_i,h_{j'})为不同语义块间负样本的均值，则该式可简化如下：
since within the same semantic block, the difference between samples, as expressed by $\bigtriangleup\text{diff}_{\text{intra}}$ in Equation (5), is much smaller than $\bigtriangleup\text{diff}_{\text{inter}}$, we assume $\theta(h_i,h_j) = 1$ for simplicity, 
% Additionally, let $\hat{\theta}(h_i,h_{j'})$ represent the mean similarity for negatives across different semantic blocks. 
% Under this assumptions, 
which can be simplified as follows:
\begin{equation}
\begin{aligned}
        % Pe^{1/\tau} &= (N-P)e^{\hat{\theta}(h_i,h_{j'})/\tau}, \\
        Pe^{1/\tau} &= \sum_{j'=1,j'\neq i}^{N-P} e^{\theta(h_i,h_{j'})/\tau}.
\end{aligned} 
\end{equation}
% 当这个实际比率大于该值时，模型会趋向于降低同一个语义块内的负样本对的相似度，而反之，模型才会降低属于不同语义块的负样本对的相似度。这说明InfoNCE目前的采样方式是存在效能阈值的，当梯度比例在该阈值上下震荡时，模型对于语义块内和语义块间的负样本，区分度就会降低，从而影响模型的性能。
% we obtain the value of $\hat{\theta}(h_i,h_{j'})$ as $1-\tau \ln((N-P)/P)$. 
When the sum of $\phi(i, j')$ 
exceeds this value, the model tends to reduce the similarity of negatives within the same semantic block. Conversely, if the sum of $\phi(i, j')$ is below this value, the model focuses on reducing the similarity of negatives across different semantic blocks. This indicates that the current sampling method in InfoNCE has an efficiency threshold. When the gradient ratio fluctuates around this threshold, the model's ability to distinguish between negatives within and across semantic blocks decreases, impacting its performance.

\textbf{Finding 1: A large number of negative nodes can hinder the model's ability to distinguish nodes with different semantics.} A large number of negatives does not improve the model's ability to differentiate between core semantics. Theorem 2 establishes the existence of a threshold for negatives. When the ratio exceeds this threshold, the model starts to distinguish the $\epsilon$ between negatives, hindering performance improvement.

\textbf{Finding 2: A smaller number of high-quality and non-topologically coupled negative nodes are sufficient to enhance the discriminability of representations.} To enhance model performance, it is crucial to differentiate the semantics of nodes belonging to distinct semantic blocks. This differentiation primarily relies on separating the core semantics. Thus, it is adequate to sample solely the nodes with core semantics as negatives, instead of sampling all nodes.

\begin{figure*}[t]
    \centering
    \includegraphics[width=0.99\textwidth]{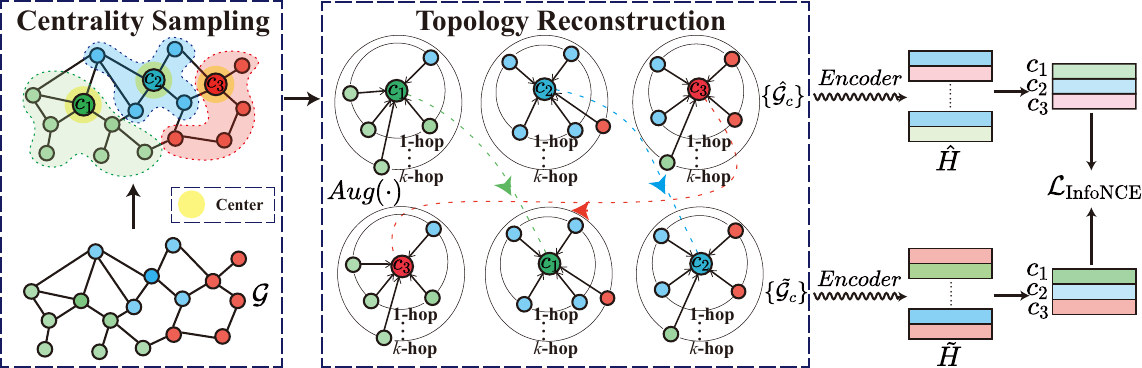}
    \caption{Given a graph $\mathcal{G}$, we use a centrality sampling strategy to select a representative set of nodes from $\mathcal{G}$. Next, we perform topological reconstruction based on the neighbors of these nodes to generate a reconstructed graph $\mathcal{\hat{G}}$. The function $Aug(\cdot)$ is a custom augmentation used to create augmented graph $\mathcal{\tilde{G}}$. $\mathcal{\hat{G}}$ and $\mathcal{\tilde{G}}$ are then input into the encoder $f$ to generate representations $\bm{\hat{H}}$ and $\bm{\tilde{H}}$. Finally, we select the representative node set to compute the loss.}
    \label{fg:overview}
\end{figure*}
%77:Methodology
\section{Methodology}
In the previous section, we obtain two key findings by taking the derivative of the InfoNCE loss about negative samples, demonstrating that negatives exhibit certain limitations during training, highlighting flaws in the existing sampling strategy. In this section, we primarily introduce the mechanism of E2Neg.
\subsubsection{Centrality Sampling.}
We calculate the normalized Laplacian matrix $\bm{L}$ and then apply EigenValue Decomposition (EVD). For the resulting eigenvectors $\bm{U}$, we select the low-frequency ones, as these vectors effectively capture the critical topological structures of the graph. 
Specifically, we select the first $k$ eigenvectors corresponding to the smallest eigenvalues $\bm{U}_k = [\bm{u}_1, \bm{u}_2, \ldots, \bm{u}_k]$. We apply the K-means \cite{Kmeans} algorithm to the rows of $\bm{U}_k$ to partition the nodes into $k$ clusters:
\begin{equation}
\mathbf{y} = \text{K-means}\left( \{\bm{u}_i \mid \bm{u}_i \in \bm{U}_k \}\right),
\end{equation}
\noindent where $\lambda_k$ is used to select the low-frequency eigenvectors. The set $\{\bm{u}_i \mid \bm{u}_i \in \bm{U}_k\}$ consists of eigenvectors with eigenvalues $ \lambda_i \leq \lambda_k$, which are then clustered using K-means.
$\mathbf{y} = \{y_1, y_2, \dots, y_n\}$ is the vector of cluster assignments for each node, with $y_i$ representing the cluster label assigned to $v_i$. For each cluster $C_j$, where $C_j = \{v_i \mid y_i = j\}$, we further analyze the intra- and inter-cluster properties. 
By evaluating the node's influence on the cluster's structural integrity, we identify a representative node, the cluster center, which is chosen to capture the cluster's spectral properties.
% based on the norm of its spectral feature as the typical representation of the cluster's spectral properties.
% which is chosen as the node that best captures the cluster's spectral properties, determined by the norm of its spectral feature vector.
These features are derived from the eigenvectors corresponding to the smallest eigenvalues of the graph's Laplacian matrix, which encapsulate the essential connectivity and structure of the cluster. The cluster center $c_i$ for cluster $C_i$ is determined by:
\begin{equation}
c_i = \arg\max_{v \in C_i} \|\mathbf{x}_v\|,
% c_i = \underset{v \in C_i}{\text{argmax}} \, \left( \|\mathbf{L}^{-1/2} \mathbf{x}_v \|_2 \right)
\end{equation}
where $\mathbf{x}_v$ represents the spectral feature of $v$. The norm $\|\mathbf{x}_v\|$ quantifies the node's centrality within the spectral domain, ensuring that the selected cluster center represents the cluster's overall structure.

\subsubsection{Topology Reconstruction.}
Following the sampling process, $\mathcal{G}$ is preprocessed into multiple clusters. For each cluster $C_i$, we select a cluster center $c_i$ based on cluster centrality. To facilitate graph reconstruction, we identify the $\hat{k}$-hop neighbors of each $c_i$ (denoted as $\hat{k}$) and establish direct connections between these neighbors and the center. This procedure results in the construction of subgraphs, each centered around its respective cluster center, as described below:
\begin{equation}
    V_i^{(\hat{k})} = \{ v_j \in \mathcal{V} \mid \text{dist}(c_i, v_j) \leq \hat{k} \} \cup \{c_i\},
\end{equation}
where the edge set of the subgraph $E_{i}^{(\hat{k})}$ is given by:
\begin{equation}
    E_i^{(\hat{k})} = \{(v_j \to c_i) \mid v_j \in V_i^{(\hat{k})} \setminus \{c_i\} \},
\end{equation}
where  $v_j \to c_i$ indicates a directed edge from $v_j$ to $c_i$. For each $c_i$, a corresponding subgraph $\mathcal{\hat{G}}_{i}$ is generated. When constructing these subgraphs, we consider the potential issue of topological coupling, which may lead to redundant information propagation, causing a node's features to be transmitted to multiple cluster centers. To address this, we ensure that each neighboring node is connected to at most one cluster center. Additionally, all subgraphs are constructed with directed edges to avoid creating redundant propagation structures, ensuring that the model focuses on training centers rather than the neighboring nodes. Thus, for the selected $k$ cluster centers, each will generate a corresponding subgraph, forming a new reconstructed graph 
$\mathcal{\hat{G}} = (\hat{\bm{A}}, \bm{X}) =  \{ \mathcal{\hat{G}}_{c} \mid c=1,2,\cdots,k\}$
, where $\hat{\bm{A}}_{i}=(V_i^{(\hat{k})}, E_i^{(\hat{k})}), \forall i \in \{1, 2, \cdots, k\}$, and $\hat{\bm{A}}_{i} \cap \hat{\bm{A}}_{j} = \varnothing, \forall i \neq j$. In the following components of the model, we replace $\mathcal{G}$ with $\mathcal{\hat{G}}$ for training.
% 对于每个簇心$c_i$都会对应生成一张子图，我们将其设为$G_i$。并且我们在构图时，考虑到可能存在的拓扑耦合问题可能会带来信息的冗余传播，从而导致一个节点的特征传递给多个簇心。因此对于每一个邻居节点，我们使其最多只与一个簇心相连。此外，为了不生成冗余的传播结构，我们对所有的子图均构建单向边，这使得模型聚焦于训练关键的簇心而不是其余的邻居节点。因此，对于筛选的k个簇心来说，每个簇心都会生成一张对应的子图，这些子图构成一个新的重构图$G_{\text{Recon}}= \{ G_{r1}, G_{r2}, \cdots, G_{rk} \}$，在后续的模型组件中，我们用$G_{\text{Recon}}$来替代$\mathcal{G}$参与模型的训练。

\subsubsection{Augmentation Strategy.}
% 对于一张重构图$G_{\text{Recon}}$, 按照传统的方法，我们可以通过Feature Masking以及Edge Droping增强这张图来生成增强图$\hat{G_{\text{Recon}}}$. 但由于图$G_{\text{Recon}}$中点和边的数量相比于原图大幅减少，故传统的增强方式只能添加少量的扰动。在这里我们使用我们自定义的增强方式：\textbf{簇心交换}。通过将重构图中的簇心之间任意交换其特征，从而生成增强图$\hat{G_{\text{Recon}}}$. 通过这种方式，在重构图$G_{\text{Recon}}$中添加扰动信息，并且$G_{\text{Recon}}$和$\hat{G_{\text{Recon}}}$会一并送入模型的编码器进行训练。$\mathcal{\tilde{G}}$ from a reconstructed graph $\mathcal{\hat{G}}$ using standard methods, we employ feature masking and edge dropping techniques. 
Traditional augmentation strategies in GCLs usually employ feature masking and edge dropping to generate the augmented graphs. Nevertheless, due to the substantial reduction in the number of nodes and edges in $\mathcal{\hat{G}}$ compared to $\mathcal{G}$, conventional augmentation strategies can only introduce a restricted amount of disruption. Therefore, we employ a specialized augmentation for $\mathcal{\hat{G}}$, which entails randomly exchanging the cluster centers in $\mathcal{\hat{G}}$ to create the augmented graph, denoted as 
% $\mathcal{\tilde{G}} = (\hat{\bm{A}}, \tilde{\bm{X}})$
$\mathcal{\tilde{G}} = (\tilde{\bm{A}}, \tilde{\bm{X}}) =  \{ \mathcal{\tilde{G}}_{c} \mid c=1,2,\cdots,k\}$
. The difference between $\mathcal{\hat{G}}$ and $\mathcal{\tilde{G}}$ can be expressed as $c_i^{\mathcal{\hat{G}}} \neq c_i^{\mathcal{\tilde{G}}}, \forall i \in \{1, 2, \cdots, k\}$, where $c_i^{\mathcal{\hat{G}}}$ and $c_i^{\mathcal{\tilde{G}}}$ represent the centers of $\mathcal{\hat{G}}_{i}$ and $\mathcal{\tilde{G}}_{i}$ respectively. Both $\mathcal{\hat{G}}$ and $\mathcal{\tilde{G}}$ are utilized to train the encoder $f$.

\subsubsection{Encoder.}
We use GNN to acquire the node representations by the node attributes and topology structure of the graph, which consists of an Encoder and a Projector. The backbone GNN Encoder may use different models (e.g., GCN \cite{GCN}, GAT \cite{GAT}). In E2Neg, we set GCN as our backbone and Multi-Layer Perception (MLP) as the Projector. The Encoder uses $\mathcal{\hat{G}}$ and $\mathcal{\tilde{G}}$ as inputs and separately obtains the reconstructed representation $\bm{\hat{H}}$ and the augmented representation $\bm{\tilde{H}}$.

\subsubsection{Loss Function.}
We adopt the InfoNCE \cite{InfoNCE} as the loss function. Unlike the settings of other GCLs, we do not feed the entire representations of $\mathcal{\hat{G}}$ and $\mathcal{\tilde{G}}$ into the loss calculation. To correspond with our findings, we optimize the model using only the representations of $c_i^{\mathcal{\hat{G}}}$ and $c_i^{\mathcal{\tilde{G}}}$
% in $\mathcal{\hat{G}}_{i}$ and $\mathcal{\tilde{G}}_{i}$
, as shown in the following equation:
\begin{equation}
\begin{aligned}&
\ell\left(\boldsymbol{\hat{h}}_{i},\boldsymbol{\tilde{h}}_{i}\right)=\\&\log\frac{e^{\theta(\boldsymbol{\hat{h}}_{i},\boldsymbol{\tilde{h}}_{i})/\tau}}{e^{\theta(\boldsymbol{\hat{h}}_{i},\boldsymbol{\tilde{h}}_{i})/\tau}+\sum_{j\neq i}^{k-1}(e^{\theta(\boldsymbol{\hat{h}}_{i},\boldsymbol{\tilde{h}}_{j})/\tau}+e^{\theta(\boldsymbol{\hat{h}}_{i},\boldsymbol{\tilde{h}}_{j})/\tau})},
\end{aligned}
\end{equation}
where $i \in \{1, 2, \cdots, k\}$. $\boldsymbol{\hat{h}}_{i}$ and $\boldsymbol{\tilde{h}}_{i}$ respectively represent the embeddings of $c_i^{\mathcal{\hat{G}}}$ and $c_i^{\mathcal{\tilde{G}}}$ in 
% $\mathcal{\hat{G}}_{i}$ and $\mathcal{\tilde{G}}_{i}$
$\bm{\hat{H}}$ and $\bm{\tilde{H}}$.

\begin{table*}[ht]
\centering
\resizebox{1.0\textwidth}{!}{
\begin{tabular}{lccccccccc}
\toprule
\textbf{Method}     & \textbf{Data}  & \textbf{PubMed} & \textbf{CS} & \textbf{Photo} & \textbf{Computers} & \textbf{Wiki-CS} & \textbf{Physics}\\ 
\midrule
Raw Features        &    $X$            & 84.80 & 90.37 & 78.53 & 73.81 & 71.98 & 93.58 \\
Node2Vec            &    $A$            & 80.30 & 85.08 & 89.67 & 84.39 & 71.79 & 91.19 \\
DeepWalk            &    $A$            & 80.50 & 84.61 & 89.44 & 85.68 & 74.35 & 91.77 \\ 
DeepWalk + Features &    $X,A$          & 83.70 & 87.70 & 90.05 & 86.28 & 77.21 & 94.90 \\ \midrule
BGRL                &    $X,A$          & 85.88 ± 0.83 & 92.16 ± 0.35 & 92.26 ± 0.92 & 87.51 ± 0.74 & 79.49 ± 0.80 & 95.04 ± 0.26 \\
MVGRL               &    $X,A$          & 85.61 ± 1.10 & 92.10 ± 0.44 & 92.68 ± 0.75 & 86.66 ± 0.91 & \underline{80.34} ± 1.01 & 95.22 ± 0.35 \\
DGI                 &    $X,A$          & 85.92 ± 0.53 & 93.12 ± 0.50 & \underline{92.75} ± 0.82 & 88.35 ± 0.68 & 80.16 ± 0.91 & \underline{95.77} ± 0.43 \\
GBT                 &    $X,A$          & 86.37 ± 1.03 & \underline{93.18} ± 0.50 & 92.39 ± 0.86 & \textbf{88.91} ± 0.98 & 80.31 ± 1.30 & \underline{95.77} ± 0.24 \\
GRACE               &    $X,A$          & 85.80 ± 0.58 & 92.70 ± 0.74 & 92.01 ± 0.85 & 88.17 ± 0.91 & 79.94 ± 0.68 & OOM \\
GCA                 &    $X,A$          & 86.44 ± 0.19 & 92.41 ± 0.08 & 91.15 ± 0.18 & 86.58 ± 0.32 & 79.87 ± 0.44 & OOM \\
ProGCL              &    $X,A$          & \underline{86.46} ± 0.54 & 92.30 ± 0.55 & 92.72 ± 1.03 & 87.65 ± 0.87 & 77.85 ± 1.06 & OOM \\
\textbf{E2Neg}      &    $X,A$          & \textbf{86.72} ± 1.09 & \textbf{93.48} ± 0.59 & \textbf{93.36} ± 0.76 & \underline{88.72} ± 0.96 & \textbf{80.89} ± 1.21 & \textbf{95.86} ± 0.30 \\ \midrule
Supervised-GCN      &    $X,A,Y$        & 84.80 & 93.03 & 92.42 & 86.51 & 77.19 & 95.65 \\
\toprule 
\end{tabular}
}
\caption{Performance on node classification. All results highlight the best with \textbf{bold} and the runner-up with an \underline{underline}. $X, A, Y$ denote the node attributes, adjacency matrix, and labels in the datasets, OOM signifies out-of-memory on RTX 3090 GPU with 24GB of memory. Data without variance are drawn from \cite{GRACE, GCA}. }
\label{tb:Classification}
\end{table*}

\subsubsection{E2Neg Effectiveness Analysis.}
We use spectral clustering to the graph and select central nodes within each cluster to serve as negative samples. This approach allows us to maximize the distinction between negatives across different semantic blocks, rather than within the same block. Even if the cluster does not perfectly align with the semantic blocks of the graph, the difference between non-aligned negatives is amplified from $\epsilon$ to the core semantics $s$ of the respective negative samples, achieving the same training effect. To align with the homophily assumption, we reconstruct the topology by subgraph connections for nodes likely belonging to the same semantic block. By linking these nodes to their neighbors, the encoder enables direct aggregation of neighborhood information, facilitating the generation of discriminative representations.

\section{Experiments}

\subsection{Experimental Setup}

\subsubsection{Datasets.}
In our experiments, we adopt six widely-used datasets, including \textit{PubMed} \cite{Dataset1}, \textit{Amazon-Photo}, \textit{Amazon-Computers}, \textit{Coauthor-CS} \cite{Dataset2}, \textit{Coauthor-Physics} and \textit{Wiki-CS} \cite{Wiki}. The statistics of the datasets are shown in the Table \ref{tb:dataset}. A detailed introduction to these datasets is in the Appendix.

% 表本来在这

\begin{table}[tb]
    \centering
    \resizebox{0.47\textwidth}{!}{
    \fontsize{14}{14}\selectfont % 设置字号为 12pt，行距为 14pt
    \begin{tabular}{lcrrrr}
        \toprule
        \text{Dataset} & \#\textit{Nodes} & \#\textit{Edges} & \#\textit{Features} & \#\textit{Classes}\\
        \midrule
        \text{PubMed}    & 19,717    & \;\;88,651& \;\;500 & \;\;3 \\
        \text{CS}        & 18,333    & 163,788   & 6,805   & 15 \\
        \text{Photo}     & \;\;7,650 & 238,163   & \;\;745 & \;\;8 \\
        \text{Computers} & 13,752    & 491,722   & \;\;767 & 10 \\
        \text{Wiki-CS}   & 11,701    & 431,726   & \;\;300 & 10 \\
        \text{Physics}   & 34,493    & 991,848   & 8,451   & \;\;5 \\
        \bottomrule
    \end{tabular}}
    \caption{Statistics of datasets used in experiments.}
    \label{tb:dataset}
\end{table}

% Baselines
\subsubsection{Baselines.}
We compare E2Neg with three types of baseline methods, including \textbf{(1)} Classical unsupervised algorithms: DeepWalk \cite{Deepwalk} and Node2Vec \cite{Node2Vec}. \textbf{(2)} Semi-supervised baselines GCN \cite{GCN}. \textbf{(3)} GCL baselines: BGRL \cite{BGRL}, MVGRL \cite{MVGRL}, DGI \cite{DGI}, GBT \cite{GBT}, GRACE \cite{GRACE}, GCA \cite{GCA}, ProGCL \cite{ProGCL}. A detailed introduction to these methods can be found in the Appendix.

\subsubsection{Implementation Details.}
We test E2Neg on the node classification task. We use a single-layer GCN as the encoder and a two-layer MLP as the projector. The settings for all of the hyperparameters can be found in the Appendix. During testing, we use the encoder to generate representations for downstream tasks. We use 10\% of the data for training the downstream classifier and the remaining 90\% for testing, and we follow the testing methods from the PyGCL \cite{PyGCL}. All methods are implemented using PyTorch Geometric framework \cite{PyG}, and all experiments are conducted on an RTX 3090 GPU with 24GB of memory and a Core i5-12400 CPU. More implementation details can be found in the Appendix.
% , and the source code at https://github.com/hedongxiao-tju/E2Neg

\subsection{Experimental Results}
\subsubsection{Node Classification.}
In Table \ref{tb:Classification}, we present the accuracy of E2Neg and other baseline models for the node classification task. Our proposed E2Neg regularly outperforms the baseline GRACE and other sampling methods such as GCA and ProGCL across all datasets, 
% . Our approach attains superior accuracy across all datasets
indicating our sampling strategy, in contrast to complete sampling, successfully mitigates the adverse effects of sample number on model performance and facilitates downstream classification more efficiently. 
% In addition, E2Neg demonstrates greater performance in comparison to other sampling methods such as GCA and ProGCL.
Although BGRL does not rely on negative sampling, topological coupling still persists and impacts the performance. This is because the model operates on the original graph, where nodes inherently exhibit topological coupling.
E2Neg does not directly resolve topological coupling, mitigating its effects by sampling within subgraphs.
This further confirms our findings that choosing a small, representative subset of nodes as samples is enough to train the model.
% and improve its capacity to differentiate nodes with distinct semantics.

\begin{table*}[ht]
\centering
\resizebox{1.0\textwidth}{!}{
\begin{tabular}{lcccccccccc}
\toprule
\textbf{Dataset}     & \textbf{ } & BGRL & MVGRL & DGI & GBT & GRACE & GCA & ProGCL & \textbf{E2Neg} & Improvement \\ 
\midrule
\textbf{PubMed}     & Mem & 7,166  & 22,466 & 970   & 1,432  & 11,610 & 11,180 & 23,054 & 404   & \textbf{58.4-98.2\%} \\
\multirowcell{2}{\textbf{ }}   & Time  & 0.0427 & 0.5412 & 0.0130 & 0.0261 & 0.2177 & 0.1598 & 0.6421 & 0.0028 & \textbf{4.6-229.3$\times$} \\
\midrule
\textbf{CS}         & Mem & 7,770  & OOM   & 2,064  & 3,324  & 11,960 & 12,660 & 22,896 & 900   & \textbf{56.4-96.1\%} \\
\multirowcell{2}{\textbf{ }}   & Time  & 0.0637 & /     & 0.0281 & 0.0597 & 0.2189 & 0.1603 & 0.5727 & 0.0030 & \textbf{9.4-190.9$\times$} \\
\midrule
\textbf{Photo}      & Mem & 2,322  & 14,690 & 1,154  & 2,156  & 3,694  & 2,478  & 5,304  & 388   & \textbf{66.4-97.4\%} \\
\multirowcell{2}{\textbf{ }}   & Time  & 0.0284 & 0.4125 & 0.0185 & 0.0273 & 0.0548 & 0.0361 & 0.1141 & 0.0028 & \textbf{6.6-147.3$\times$} \\
\midrule
\textbf{Computers}  & Mem & 5,722  & OOM   & 1,914  & 2,570  & 8,952  & 6,548  & 15,506 & 406   & \textbf{78.8-97.4\%} \\
\multirowcell{2}{\textbf{ }}   & Time  & 0.0566 & /     & 0.0354 & 0.0496 & 0.1382 & 0.0964 & 0.3321 & 0.0028 & \textbf{12.6-118.6$\times$} \\
\midrule
\textbf{Wiki-CS}    & Mem & 3,928  & 17,964 & 1,696  & 3,074  & 5,698  & 4,898  & 11,402 & 380   & \textbf{77.6-97.9\%} \\
\multirowcell{2}{\textbf{ }}   & Time  & 0.0465 & 0.5139 & 0.0300 & 0.0418 & 0.1073 & 0.0731 & 0.2470 & 0.0028 & \textbf{10.7-183.5$\times$} \\
\midrule
\textbf{Physics}    & Mem & 23,688 & OOM   & 4,218  & 8,002  & OOM   & OOM   & OOM   & 1,560  & \textbf{63.0-93.4\%} \\
\multirowcell{2}{\textbf{ }}   & Time  & 0.1635 & /     & 0.0675 & 0.1319 & /      & /      & /      & 0.0031 & \textbf{21.8-52.7$\times$} \\
\bottomrule 
\end{tabular}
}
\caption{Comparison of training time per epoch in \underline{seconds} and memory usage in \underline{MBs} on six datasets. Improvement means how many times E2Neg is faster than baselines or the percentage of memory savings that E2Neg provides compared to baselines. OOM signifies out-of-memory on 24GB RTX 3090. '-’ means the improvement range.}
\label{tb:Efficiency}
\end{table*}

\begin{figure*}[ht]
\centering
\subfigure{
\includegraphics[width=0.23\linewidth]{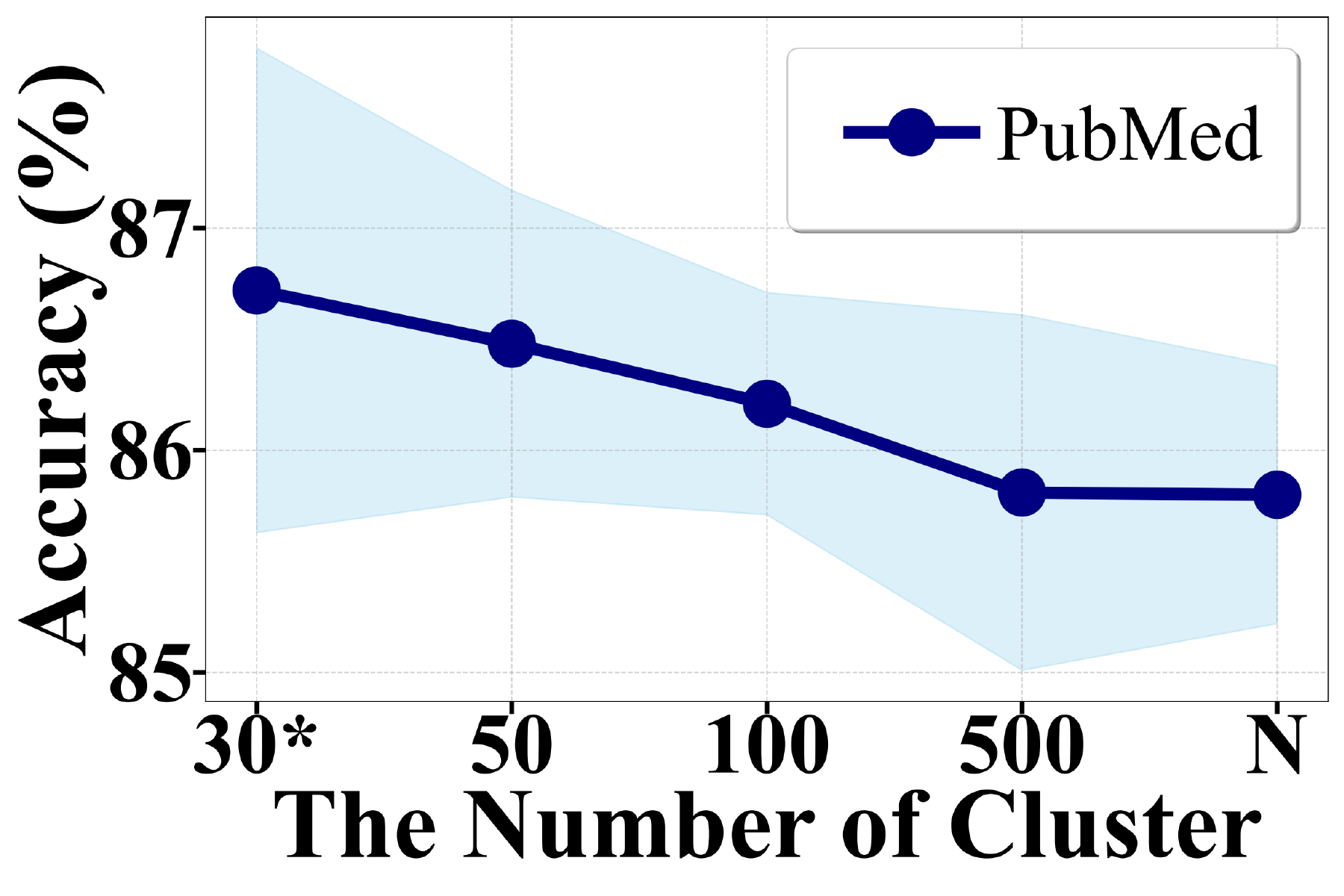}
}
\subfigure{
\includegraphics[width=0.23\linewidth]{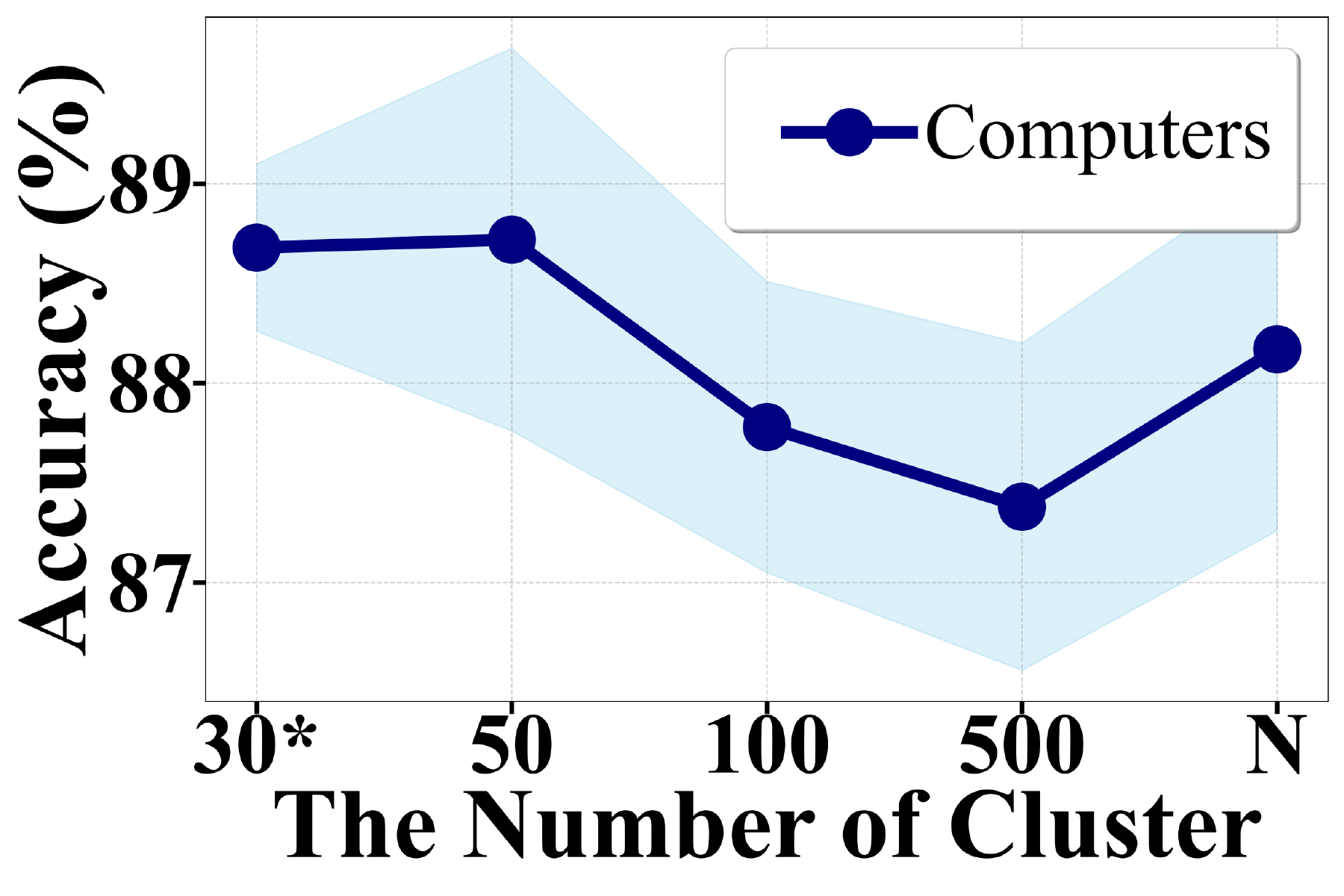}
}
\subfigure{
\includegraphics[width=0.23\linewidth]{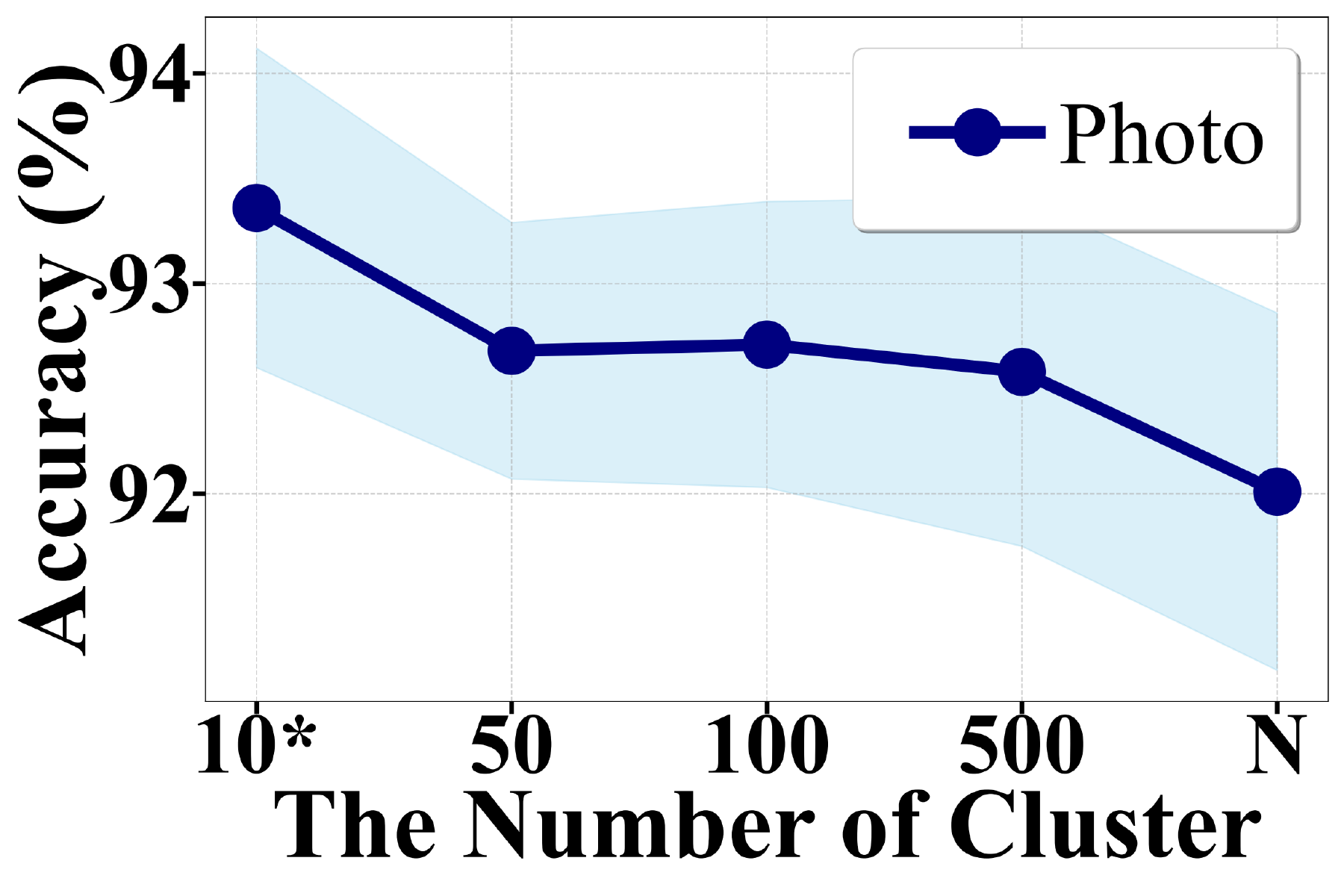}
}
\subfigure{
\includegraphics[width=0.23\linewidth]{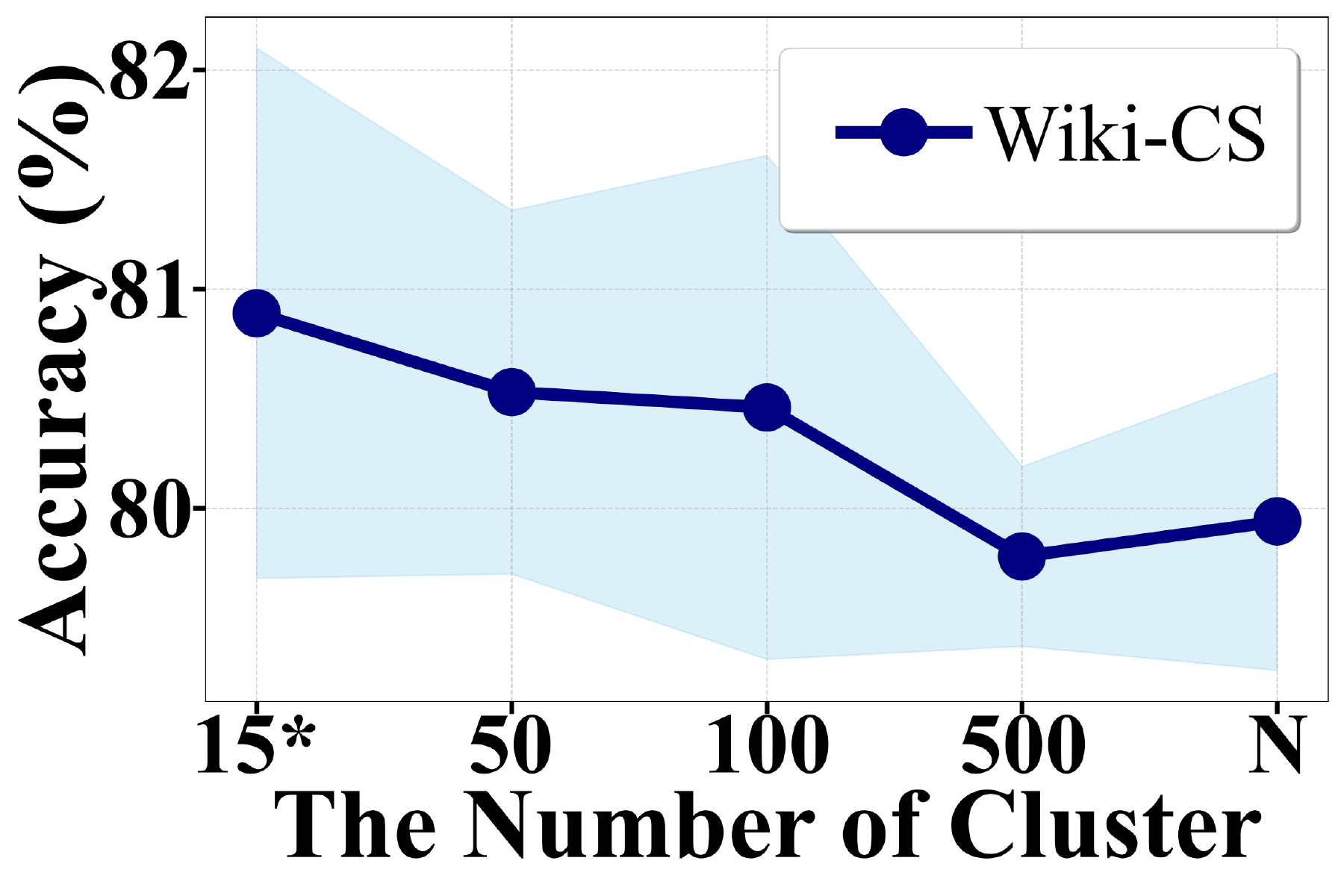}
}
\caption{Hyperparameter Analysis, where * denotes the parameter used by E2Neg, and N represents the number of nodes in each dataset.}
\label{tb:hyperparameter}
\end{figure*}

\subsubsection{Computational Complexity Analysis.}
E2Neg demonstrates significantly higher efficiency in both time and memory consumption compared to other self-supervised baselines, as shown in Table \ref{tb:Efficiency}. For a fair comparison, we set the number of hidden dimension to 256 for all methods, with other parameters consistent with those in Table \ref{tb:Classification}. Its strategy of selecting representative nodes for training reduces memory utilization by over 90\% on average compared to sampling-based baselines like GCA and ProGCL, which have high memory overhead. In terms of training time, E2Neg achieve remarkable improvement, with average speedups ranging from tens to hundreds of times. On the larger dataset, \textit{Physics}, E2Neg is over 20 times faster than all other methods, and it consistently achieves speedup factors exceeding 100 times across all datasets.
% 这段一般
Moreover, it is evident that traditional sampling methods, such as GRACE, GCA, and ProGCL, consume a significant amount of memory and time across these six datasets. Additionally, for other types of methods like BGRL and DGI, the traditional sampling approaches do not offer competitive efficiency, whereas our E2Neg significantly outperforms these methods in terms of efficiency. The significant improvement in time and memory efficiency of E2Neg can be attributed to its substantial reduction in the sample, which prevents the model from generating a large number of negatives for similarity calculations during training.

\begin{table}[ht]
\centering
\resizebox{0.48\textwidth}{!}{
\begin{tabular}{lccccc}
\toprule
\textbf{Method}     & Random  & Full  & No Aug & E2Neg \\ 
\midrule
CS         & 92.87 ± 0.47  & 92.70 ± 0.74  & 93.12 ± 0.59  & \textbf{93.48} ± 0.59 \\
\midrule
Photo      & 92.42 ± 1.31 & 92.01 ± 0.85  & 92.54 ± 0.96  & \textbf{93.36} ± 0.76 \\
\midrule
Computers  & 87.83 ± 0.70 & 88.17 ± 0.91  & 88.18 ± 0.74  & \textbf{88.72} ± 0.96 \\
\midrule
Wiki-CS    & 79.80 ± 1.22 & 79.94 ± 0.68  & 79.45 ± 1.12  & \textbf{80.89} ± 1.21 \\
\midrule
Physics    & 94.97 ± 0.32 & OOM           & 94.53 ± 0.45  & \textbf{95.86} ± 0.30 \\
\bottomrule 
\end{tabular}
}
\caption{Ablation study with several components of E2Neg. All results highlight the best with bold.}
\label{tb:Ablation}
\end{table}

\subsection{Model Analysis}
\subsubsection{Ablation Study.}
To further validate the correctness of our theory and the effectiveness of E2Neg. In this section, we substitute several components of E2Neg to analyze the impact of each component. In our study, we modify the sampling strategy by using random and full sampling. Additionally, we remove the augmentation from E2Neg. The performance is presented in Table \ref{tb:Ablation}. E2Neg outperforms all the different experimental variations. For the sampling strategy, E2Neg selects representative nodes that are more effective in capturing the core semantics of the graph. Compared to random sampling, these representative nodes significantly enhance the model's ability to distinguish negative samples across different semantic blocks, leading to improved discriminative performance. E2Neg provides a more representative sample than selecting all nodes, which aligns with Finding 1 from our theoretical analysis. The augmentation enables individual samples to acquire the semantic information available in various subgraphs, improving the discriminative capability of the representations.
% xxxxxxxxxxxxxxxxxxxxxxxxxxxxxxxxxxxxxxxxx
% xxxxxxxxxxxxxxxxxxxxxxxxxxxxxxxxxxxxxxxxx

\subsubsection{Hyperparameter Analysis.}
To further explore the correctness and effectiveness of the E2Neg sampling and to substantiate the analysis in our theoretical analysis, we test the impact of the number of clusters on the model's performance, as shown in Figure \ref{tb:hyperparameter}. It can be observed that when the cluster size is large, the performance of E2Neg decreases across various datasets. This is because the number of clusters far exceeds the actual number of semantic blocks in the graph, which declines the model's ability to distinguish between negatives from different semantic blocks. Conversely, keeping the cluster size small allows the number of clusters to align with the semantic blocks, leading to an improvement in model performance. This is consistent with our findings in the theoretical analysis section, where we observe that selecting a representative subset of nodes, 
% rather than a large number of nodes, 
significantly enhances the model's performance. The experimental results further validate the correctness of our theoretical analysis and demonstrate the effectiveness of E2Neg.

\section{Conclusion}
In this paper, we explore the role of negative samples in GCL. Specifically, we challenge the consensus of selecting all instances as negative samples and theoretically discover that a large number of negatives can hinder the model's ability to distinguish nodes with different semantics. Furthermore, using only a very small subset of nodes as negatives is sufficient to enhance the discriminative power of the representations. Based on these findings, we propose E2Neg, which trains using only a very small set of representative negatives. E2Neg reduces computational overhead and significantly accelerates model training. Finally, we conduct extensive experiments on multiple datasets. The experimental results demonstrate the effectiveness of E2Neg.

\section{Acknowledgments}
This work was supported by the National Natural Science Foundation of China (No. U22B2036, No. 62422210 , No. 62276187, No. 92370111 and No. 62272340), the National Science Fund for Distinguished Young Scholarship (No. 62025602), and the XPLORER PRIZE.

\bibliography{aaai25}

\begin{thebibliography}{28}
\providecommand{\natexlab}[1]{#1}

\bibitem[{Bielak, Kajdanowicz, and Chawla(2022)}]{GBT}
Bielak, P.; Kajdanowicz, T.; and Chawla, N.~V. 2022.
\newblock Graph Barlow Twins: {A} self-supervised representation learning framework for graphs.
\newblock \emph{Knowl. Based Syst.}, 256: 109631.

\bibitem[{Chen et~al.(2020)Chen, Kornblith, Norouzi, and Hinton}]{SimCLR}
Chen, T.; Kornblith, S.; Norouzi, M.; and Hinton, G.~E. 2020.
\newblock A Simple Framework for Contrastive Learning of Visual Representations.
\newblock In \emph{Proceedings of the 37th International Conference on Machine Learning}, 1597--1607.

\bibitem[{Fan et~al.(2019)Fan, Ma, Li, He, Zhao, Tang, and Yin}]{SocialRecommendation}
Fan, W.; Ma, Y.; Li, Q.; He, Y.; Zhao, Y.~E.; Tang, J.; and Yin, D. 2019.
\newblock Graph Neural Networks for Social Recommendation.
\newblock In \emph{The World Wide Web Conference}, 417--426.

\bibitem[{Fey and Lenssen(2019)}]{PyG}
Fey, M.; and Lenssen, J.~E. 2019.
\newblock Fast Graph Representation Learning with PyTorch Geometric.
\newblock \emph{CoRR}, abs/1903.02428.

\bibitem[{Grover and Leskovec(2016)}]{Node2Vec}
Grover, A.; and Leskovec, J. 2016.
\newblock node2vec: Scalable Feature Learning for Networks.
\newblock In \emph{Proceedings of the 22nd {ACM} {SIGKDD} International Conference on Knowledge Discovery and Data Mining}, 855--864.

\bibitem[{Hamilton, Ying, and Leskovec(2017)}]{GraphSAGE}
Hamilton, W.~L.; Ying, Z.; and Leskovec, J. 2017.
\newblock Inductive Representation Learning on Large Graphs.
\newblock In \emph{Advances in Neural Information Processing Systems 30}, 1024--1034.

\bibitem[{Hassani and Ahmadi(2020)}]{MVGRL}
Hassani, K.; and Ahmadi, A. H.~K. 2020.
\newblock Contrastive Multi-View Representation Learning on Graphs.
\newblock In \emph{Proceedings of the 37th International Conference on Machine Learning}, 4116--4126.

\bibitem[{He et~al.(2020)He, Fan, Wu, Xie, and Girshick}]{MoCo}
He, K.; Fan, H.; Wu, Y.; Xie, S.; and Girshick, R.~B. 2020.
\newblock Momentum Contrast for Unsupervised Visual Representation Learning.
\newblock In \emph{{IEEE/CVF} Conference on Computer Vision and Pattern Recognition}, 9726--9735.

\bibitem[{Jin et~al.(2021)Jin, Wang, He, Dang, and Zhang}]{JinW21}
Jin, D.; Wang, X.; He, D.; Dang, J.; and Zhang, W. 2021.
\newblock Robust Detection of Link Communities With Summary Description in Social Networks.
\newblock \emph{{IEEE} Trans. Knowl. Data Eng.}, 33(6): 2737--2749.

\bibitem[{Jin et~al.(2023)Jin, Yu, Jiao, Pan, He, Wu, Yu, and Zhang}]{CommunityDetectionSurvey}
Jin, D.; Yu, Z.; Jiao, P.; Pan, S.; He, D.; Wu, J.; Yu, P.~S.; and Zhang, W. 2023.
\newblock A Survey of Community Detection Approaches: From Statistical Modeling to Deep Learning.
\newblock \emph{{IEEE} Trans. Knowl. Data Eng.}, 35(2): 1149--1170.

\bibitem[{Kipf and Welling(2017)}]{GCN}
Kipf, T.~N.; and Welling, M. 2017.
\newblock Semi-Supervised Classification with Graph Convolutional Networks.
\newblock In \emph{5th International Conference on Learning Representations}.

\bibitem[{Li et~al.(2023)Li, Wang, Xiong, and Lai}]{HomoGCL}
Li, W.; Wang, C.; Xiong, H.; and Lai, J. 2023.
\newblock HomoGCL: Rethinking Homophily in Graph Contrastive Learning.
\newblock In \emph{Proceedings of the 29th {ACM} {SIGKDD} Conference on Knowledge Discovery and Data Mining}, 1341--1352.

\bibitem[{Liu et~al.(2023)Liu, Lin, Liu, Liu, Zheng, and Dong}]{B2-sampling}
Liu, M.; Lin, Y.; Liu, J.; Liu, B.; Zheng, Q.; and Dong, J.~S. 2023.
\newblock B\({}^{\mbox{2}}\)-Sampling: Fusing Balanced and Biased Sampling for Graph Contrastive Learning.
\newblock In \emph{Proceedings of the 29th {ACM} {SIGKDD} Conference on Knowledge Discovery and Data Mining}, 1489--1500.

\bibitem[{MacQueen(1967)}]{Kmeans}
MacQueen, J. 1967.
\newblock Some methods for classification and analysis of multivariate observations.
\newblock In \emph{Proceedings of 5-th Berkeley Symposium on Mathematical Statistics and Probability/University of California Press}.

\bibitem[{Mernyei and Cangea(2020)}]{Wiki}
Mernyei, P.; and Cangea, C. 2020.
\newblock Wiki-CS: {A} Wikipedia-Based Benchmark for Graph Neural Networks.
\newblock \emph{CoRR}, abs/2007.02901.

\bibitem[{Perozzi, Al{-}Rfou, and Skiena(2014)}]{Deepwalk}
Perozzi, B.; Al{-}Rfou, R.; and Skiena, S. 2014.
\newblock DeepWalk: online learning of social representations.
\newblock In \emph{The 20th {ACM} {SIGKDD} International Conference on Knowledge Discovery and Data Mining}, 701--710.

\bibitem[{Shchur et~al.(2018)Shchur, Mumme, Bojchevski, and G{\"{u}}nnemann}]{Dataset2}
Shchur, O.; Mumme, M.; Bojchevski, A.; and G{\"{u}}nnemann, S. 2018.
\newblock Pitfalls of Graph Neural Network Evaluation.
\newblock \emph{CoRR}, abs/1811.05868.

\bibitem[{Thakoor et~al.(2021)Thakoor, Tallec, Azar, Munos, Veli{\v{c}}kovi{\'c}, and Valko}]{BGRL}
Thakoor, S.; Tallec, C.; Azar, M.~G.; Munos, R.; Veli{\v{c}}kovi{\'c}, P.; and Valko, M. 2021.
\newblock Bootstrapped representation learning on graphs.
\newblock In \emph{ICLR 2021 Workshop on Geometrical and Topological Representation Learning}.

\bibitem[{van~den Oord, Li, and Vinyals(2018)}]{InfoNCE}
van~den Oord, A.; Li, Y.; and Vinyals, O. 2018.
\newblock Representation Learning with Contrastive Predictive Coding.
\newblock \emph{CoRR}, abs/1807.03748.

\bibitem[{Veli{\v{c}}kovi{\'{c}} et~al.(2018)Veli{\v{c}}kovi{\'{c}}, Cucurull, Casanova, Romero, Li{\`{o}}, and Bengio}]{GAT}
Veli{\v{c}}kovi{\'{c}}, P.; Cucurull, G.; Casanova, A.; Romero, A.; Li{\`{o}}, P.; and Bengio, Y. 2018.
\newblock {Graph Attention Networks}.
\newblock \emph{International Conference on Learning Representations}.
\newblock Accepted as poster.

\bibitem[{Velickovic et~al.(2019)Velickovic, Fedus, Hamilton, Li{\`{o}}, Bengio, and Hjelm}]{DGI}
Velickovic, P.; Fedus, W.; Hamilton, W.~L.; Li{\`{o}}, P.; Bengio, Y.; and Hjelm, R.~D. 2019.
\newblock Deep Graph Infomax.
\newblock In \emph{7th International Conference on Learning Representations}.

\bibitem[{Wang et~al.(2023)Wang, Dong, Jin, Li, Wang, and Dang}]{Wang23}
Wang, X.; Dong, Y.; Jin, D.; Li, Y.; Wang, L.; and Dang, J. 2023.
\newblock Augmenting Affective Dependency Graph via Iterative Incongruity Graph Learning for Sarcasm Detection.
\newblock In \emph{Thirty-Seventh {AAAI} Conference on Artificial Intelligence, {AAAI} 2023, Washington, DC, USA, February 7-14, 2023}, 4702--4710. {AAAI} Press.

\bibitem[{Xia et~al.(2022)Xia, Wu, Wang, Chen, and Li}]{ProGCL}
Xia, J.; Wu, L.; Wang, G.; Chen, J.; and Li, S.~Z. 2022.
\newblock ProGCL: Rethinking Hard Negative Mining in Graph Contrastive Learning.
\newblock In \emph{International Conference on Machine Learning}, 24332--24346.

\bibitem[{Yang, Cohen, and Salakhutdinov(2016)}]{Dataset1}
Yang, Z.; Cohen, W.~W.; and Salakhutdinov, R. 2016.
\newblock Revisiting Semi-Supervised Learning with Graph Embeddings.
\newblock In \emph{Proceedings of the 33nd International Conference on Machine Learning}.

\bibitem[{Zhang et~al.(2022)Zhang, Wu, Wang, Zhang, Yan, and Yu}]{LocalGCL}
Zhang, H.; Wu, Q.; Wang, Y.; Zhang, S.; Yan, J.; and Yu, P.~S. 2022.
\newblock Localized Contrastive Learning on Graphs.
\newblock \emph{CoRR}, abs/2212.04604.

\bibitem[{Zhu et~al.(2021{\natexlab{a}})Zhu, Xu, Liu, and Wu}]{PyGCL}
Zhu, Y.; Xu, Y.; Liu, Q.; and Wu, S. 2021{\natexlab{a}}.
\newblock An Empirical Study of Graph Contrastive Learning.
\newblock In \emph{Proceedings of the Neural Information Processing Systems Track on Datasets and Benchmarks 1, December 2021, virtual}.

\bibitem[{Zhu et~al.(2020)Zhu, Xu, Yu, Liu, Wu, and Wang}]{GRACE}
Zhu, Y.; Xu, Y.; Yu, F.; Liu, Q.; Wu, S.; and Wang, L. 2020.
\newblock Deep Graph Contrastive Representation Learning.
\newblock \emph{CoRR}, abs/2006.04131.

\bibitem[{Zhu et~al.(2021{\natexlab{b}})Zhu, Xu, Yu, Liu, Wu, and Wang}]{GCA}
Zhu, Y.; Xu, Y.; Yu, F.; Liu, Q.; Wu, S.; and Wang, L. 2021{\natexlab{b}}.
\newblock Graph Contrastive Learning with Adaptive Augmentation.
\newblock In \emph{{WWW} '21: The Web Conference}, 2069--2080.

\end{thebibliography}

\end{document}

\maketitle

\section{Related Work}

\subsubsection{Graph Neural Networks.} Graph Neural Networks (GNNs) have become a cornerstone in learning from graph-structured data, thanks to their ability to capture complex relational dependencies between nodes. The foundational work in this area, such as Graph Convolutional Networks (GCNs) \cite{GCN}, applies convolution operations to graph domains. GCNs aggregate information from a node's local neighborhood to learn effective node embeddings, making them particularly suited for tasks on homophilic graphs. However, the assumption of homophily and the risk of over-smoothing with deeper layers limit their expressiveness, particularly in complex graph structures.

To overcome some of these limitations, GraphSAGE \cite{GraphSAGE} utilizes a sampling and aggregation strategy to create node representations. By sampling a fixed number of neighbors and aggregating their features. GraphSAGE also enables the model to handle unseen nodes effectively, addressing some of the scalability issues inherent in GCNs. Graph Attention Networks (GATs) \cite{GAT} marked another significant advancement by incorporating attention mechanisms into the aggregation process. which allow the model to assign different importance weights to different neighbors, enabling more fine-grained and context-aware representation learning. While these models have made significant strides in graph representation learning, they primarily focus on local neighborhood aggregation, often overlooking the broader topological structure of the graph. Unlike traditional methods that rely on fixed or learnable neighborhood aggregation, topological receptive fields consider the global topological context, providing a more holistic understanding of a node's position and role within the graph. These approaches aim to capture the nuanced topological features of graphs, enabling models to better understand and leverage the underlying structure, particularly in scenarios where local neighborhood information alone is insufficient.

\subsubsection{Graph Contrastive Learning.} Graph Contrastive Learning (GCL) currently has attracted widespread attention in the academic community. It mainly generates multiple augmented views through augmentation, and designs objective functions to train the model based on maximizing mutual information to reduce the model's dependence on label information. As a classic paradigm, GRACE \cite{GRACE} trains the model by maximizing the similarity of nodes at the same position in two views and minimizing the similarity of nodes at other positions. On this basis, GCA \cite{GCA} designed an adaptive enhanced GCL framework to measure the importance of nodes and edges to protect the semantic information of graph data during augmentation. Local-GCL \cite{LocalGCL} treats first-order neighbors as positive samples and employs a kernel-based contrastive loss to reduce complexity. ProGCL \cite{ProGCL} utilizes a Beta mixture model to estimate the probability of a negative sample being true and devises methods to compute the weight of negative samples and synthesize new negative samples. $\text{B}^{2}$-Sampling \cite{B2-sampling} employs a balanced sampling strategy to select negative samples and corrects the noisy labels within these samples. HomoGCL \cite{HomoGCL} expands the number of positive sample pairs using a Gaussian mixture model (GMM) and calculates the weight of positive samples through soft clustering. BGRL \cite{BGRL} adopts BYOL \cite{BYOL} as the backbone, where the online encoder is trained by predicting the target encoder to generate efficient node representations. Like BGRL, AFGRL \cite{AFGRL} also employs BYOL as the backbone to eliminate the dependency on negative samples. AFGRL determines positive samples using K Nearest Neighbor (KNN) and clustering algorithms. Unlike the GCLs mentioned above, DGI \cite{DGI} learns embeddings by maximizing the mutual information between node representations and graph representations, while GGD \cite{GGD} divides nodes into positive and negative groups and uses simple binary cross-entropy to distinguish between these groups. MVGRL \cite{MVGRL} generates a global view using a diffusion matrix and contrasts it with a local view, combining global and local information.

% % GNN（要引到拓扑感受野，大多数基于传播聚合、总结拓扑感受野）、GCL（GCL是啥，有几个范式，在相关工作，在GCL下的方法DGI、BGRL，数据增强扩散MVGRL）、NegativeSampling（自己搜一搜）
\subsubsection{Negative Sampling.} Negative sampling plays a pivotal role in contrastive learning frameworks, where the objective is to distinguish between similar and dissimilar pairs. This concept builds on Noise Contrastive Estimation (NCE) \cite{NCE} , where the task is to differentiate true data points from noise introduced into the dataset. In NCE, negative samples are drawn from a noise distribution and are essential for the model to learn meaningful representations by contrasting them with positive samples.

In the field of visual representation learning, several landmark studies have emphasized the importance of negative sampling. MoCo \cite{MoCo} introduces a dynamic memory bank that stores a large set of negative samples, which are then contrasted with positive pairs to learn robust embeddings. The approach allows for a more flexible and efficient retrieval of negative samples, which is critical in maintaining the quality of the learned representations. SimCLR \cite{SimCLR} demonstrates that using larger batch sizes to generate numerous negative pairs during training significantly enhances performance. Both MoCo and SimCLR operate on the principle that the more negative samples a model is exposed to, the better it can distinguish between positive and negative instances, leading to more discriminative representations.

Further advancements in negative sampling strategies have been explored in methods like InfoNCE \cite{InfoNCE} and PIRL \cite{PIRL}. InfoNCE is designed to optimize the model by comparing each positive sample against multiple negatives. PIRL extends this by leveraging negative samples that are harder to distinguish from positives, thereby pushing the model to learn more nuanced representations.

\section{Theoretical Supplement}
In order to make the readers more clear about the motivation of our article, we provide the proof of our Theorem here.

\subsection{Proof of Theorem 1}
% The proof of Semantic Block-Based Decomposition: 
\begin{proof}
We assume the graph contains $k$ semantic blocks $\mathcal{S}=\{\mathcal{S}_{1}, \dots, \mathcal{S}_{k}\}$, each with a core semantic $s_j$. For an anchor node $v_i$ belonging to semantic block $\mathcal{S}_j$, its feature can be decomposed as:
\begin{equation}
    x_i = s_j + \epsilon_i,
\end{equation}
where $s_j$ is the core semantic of block $\mathcal{S}_j$, and $\epsilon_i$ represents the individual deviation of node $v_i$.

\textbf{Intra-Block Difference.}
For two nodes $v_i$ and $v_j$ within the same semantic block $\mathcal{S}_p$, their features are:
\begin{equation}
    x_i = s_p + \epsilon_i, \quad x_j = s_p + \epsilon_j,
\end{equation}
and the difference between their features is:
\begin{equation}
    \Delta \text{diff}_{\text{intra}} = \| x_i - x_j \| = \| \epsilon_i - \epsilon_j \|,
\end{equation}
indicating that the intra-block difference is determined by the individual deviations from the core semantic.

\textbf{Inter-Block Difference.}
For two nodes $v_i$ and $v_j$ from different semantic blocks $\mathcal{S}_p$ and $\mathcal{S}_q$, their features are:
\begin{equation}
    x_i = s_p + \epsilon_i, \quad x_j = s_q + \epsilon_j,
\end{equation}
and the difference between their features is:
\begin{equation}
    \Delta \text{diff}_{\text{inter}} = \| (s_p + \epsilon_i) - (s_q + \epsilon_j) \|.
\end{equation}
Using the triangle inequality, this can be simplified to:
\begin{equation}
    \Delta \text{diff}_{\text{inter}} = \| s_p - s_q + (\epsilon_i - \epsilon_j) \| \geq \| s_p - s_q \| - \| \epsilon_i - \epsilon_j \|.
\end{equation}
When $\| s_p - s_q \|$ is sufficiently large, the individual deviation $\| \epsilon_i - \epsilon_j \|$ can be neglected, leading to:
\begin{equation}
    \Delta \text{diff}_{\text{inter}} \approx \| s_p - s_q \|,
\end{equation}
indicating that the inter-block difference is primarily determined by the difference between the core semantics of the blocks.
\end{proof}

\subsection{Proof of Theorem 2}

\begin{proof}
Consider the InfoNCE loss function with respect to the similarity between negative samples:
\begin{equation}
\frac{\partial\mathcal{L}_{\text{InfoNCE}}}{\partial \theta(h_i,h_{j})} = 
\frac{1}{\tau}
\frac{e^{\theta(h_i,h_j)/\tau}}{e^{\theta(h_i,h_i')/\tau} + \sum_{k \neq i} e^{\theta(h_i,h_k)/\tau}},
\end{equation}
where $\theta(h_i,h_j)$ represents the similarity between any pair of negative samples. For a given anchor node $v_i$, the sum of gradients for intra-semantic block negatives $\text{SG}(i)_{\text{intra}}$ and inter-semantic block negatives $\text{SG}(i)_{\text{inter}}$ can be expressed as:
\begin{equation}
\begin{aligned}
    \text{SG}(i)_{\text{intra}} &= \sum_{j \in \mathbf{N}_{\text{intra}}} \frac{\partial\mathcal{L}_{\text{InfoNCE}}}{\partial \theta(h_i,h_j)}, \quad \\
\text{SG}(i)_{\text{inter}} &= \sum_{j' \in \mathbf{N}_{\text{inter}}} \frac{\partial\mathcal{L}_{\text{InfoNCE}}}{\partial \theta(h_i,h_{j'})}.
\end{aligned}
\end{equation}

To achieve balance, we assume $\text{SG}(i)_{\text{intra}} = \text{SG}(i)_{\text{inter}}$:

\begin{equation}
\sum_{j \in \mathbf{N}_{\text{intra}}} e^{\theta(h_i,h_j)/\tau} = \sum_{j' \in \mathbf{N}_{\text{inter}}} e^{\theta(h_i,h_{j'})/\tau}.
\end{equation}

Assuming $\theta(h_i,h_j) = 1$ for intra-block pairs and letting $\hat{\theta}(h_i,h_{j'})$ denote the mean similarity for inter-block pairs, we get:

\begin{equation}
Pe^{1/\tau} = \sum_{j'=1,j'\neq i}^{N-P} e^{\theta(h_i,h_{j'})/\tau}.
\end{equation}

% When the sum of $\phi(i, j')$ 
% exceeds this value, the model tends to reduce the similarity of negatives within the same semantic block. Conversely, if the sum of $\phi(i, j')$ is below this value, the model focuses on reducing the similarity of negatives across different semantic blocks.

This defines the threshold at which the model balances its focus between distinguishing intra- and inter-semantic block negatives.
\end{proof}

\begin{table*}[h]
    \centering
    \caption{Hyperparameters specifications}
    \label{tab:hyperparameters}
    \resizebox{0.95\textwidth}{!}{
    \begin{tabular}{lcccccc}
        \toprule
        Dataset & Learning rate & Weight decay & Hidden\_dim & Num epoch & Cluster & Neighbors \\
        \midrule
        PubMed & 0.00005 & 0.0005 & 4096 & 1500 & 30 & 100 \\
        CS & 0.0001 & 0.00005 & 2048 & 1500 & 50 & 100 \\
        Photo & 0.00001 & 0.00001 & 4096 & 600 & 10 & 100 \\
        Computers & 0.00005 & 0.00001 & 4096 & 200 & 30 & 100 \\
        Physics & 0.00001 & 0.00005 & 2048 & 600 & 15 & 100 \\
        Wiki-CS & 0.00001 & 0.00005 & 512 & 200 & 15 & 10 \\
        \bottomrule
    \end{tabular}
    }
\end{table*}

\begin{table*}[h]
    \centering
    \caption{Code links of various baseline methods.}
    \label{tab:baseline}
    \resizebox{0.62\textwidth}{!}{
    \begin{tabular}{lccc}
        \toprule
        Methods & Source Code \\
        \midrule
        BGRL & https://github.com/nerdslab/bgrl  \\
        MVGRL & https://github.com/kavehhassani/mvgrl  \\
        DGI & https://github.com/PetarV-/DGI  \\
        GBT & https://github.com/pbielak/graph-barlow-twins  \\
        GRACE & https://github.com/CRIPAC-DIG/GRACE  \\
        GCA & https://github.com/CRIPAC-DIG/GCA  \\
        ProGCL & https://github.com/junxia97/ProGCL \\
        
        \bottomrule
    \end{tabular}
    }
\end{table*}

\section{Datasets}

In our experiments, we adopt six widely-used datasets, including \textit{PubMed} \cite{CitationNetwork}, \textit{Amazon-Photo}, \textit{Amazon-Computers} \cite{Amazon}, \textit{Coauthor-CS}, \textit{Coauthor-Physics} \cite{CoAuthor} and \textit{Wiki-CS} \cite{Wiki}. The detailed introduction of these datasets are as follows:

\begin{itemize}
    \item 
    \textbf{PubMed} \cite{CitationNetwork} is citation network datasets where nodes mean papers and edges mean citation relationships. Each dimension of the feature corresponds to a word. The  nodes are labeled by the categories of the paper.
    \item 
    \textbf{Amazon-Photo and Amazon-Computers}\cite{Amazon} are two networks based on Amazon co-purchase graphs, where nodes mean goods and edges mean that two goods are frequently bought together. Node features are bag-of-words vector generated from product reviews, and the nodes are labeled by product categories.
    \item 
    \textbf{Coauthor-CS and Coauthor-Physics} \cite{CoAuthor} are two co-authorship networks based on the Microsoft Academic Graph from the KDD Cup 2016 challenge, where nodes mean authors and edged mean co-authorship between two authors. Node features are paper keywords of each author’s papers, and class labels correspond to the most active fields of each author.
    \item 
    \textbf{Wiki-CS} \cite{Wiki} is a Wikipedia-based dataset where nodes mean Computer Science articles and edges mean hyperlinks between the articles. Node features are calculated as the average of pre-trained GloVe word embeddings \cite{Glove} and class labels are different branches of the field.
\end{itemize}

\section{Hyperparameters Settings}

In this section, we present the hyperparameter specifications used for training the E2Neg model on various datasets. Table \ref{tab:hyperparameters} details the hyperparameters employed for different datasets.

\section{Pseudo Code of E2Neg}

The following pseudo code outlines the E2Neg training algorithm, which combines spectral clustering with graph augmentation to improve negative sampling in GCL. The algorithm clusters nodes, identifies cluster centers, and reconstructs the graph topology. During training, it computes embeddings for both the original and augmented graphs, and updates the model using the InfoNCE loss, focusing on the cluster centers to train the embeddings.

\begin{algorithm}[tb]
\caption{The E2Neg training algorithm}
\label{alg:e2neg}
\textbf{Input}: Original Graph $\mathcal{G}$, Encoder $f$, Projector $g$.\\
\textbf{Parameter}: parameters of various datasets in Table \ref{tab:hyperparameters}. 
% \textbf{Output}: embedding $\bm{H}$, trained encoder $f$ .
\begin{algorithmic}[1] %[1] enables line numbers
\STATE Get cluster $C = \{ C_1, C_2, \cdots, C_K \}$ using K-means
\STATE Get cluster center $c = \{ c_1, c_2, \cdots, c_k \}$ via spectual centrality
\STATE Get $\hat{\mathcal{G}}$ by topology reconstruction of $\mathcal{G}$
\STATE \textbf{for} epoch = 0, 1, 2, $\ldots$ \textbf{do}

    \STATE \hspace{1em} Generate an augmentation function $t$
    \STATE \hspace{1em} Get augmented graph via $\Tilde{\mathcal{G}} = t(\hat{\mathcal{G}})$
    \STATE \hspace{1em} Get node embedding $\bm{\hat{H}}$, $\bm{\tilde{H}}$ of $\hat{\mathcal{G}}$, $\Tilde{\mathcal{G}}$ through the same encoder $f$.
    \STATE \hspace{1em} Compute InfoNCE loss $\mathcal{L}$ only using the representations of cluster centers.
    \STATE \hspace{1em} Update the parameters of $f$ via $\mathcal{L}$ 
\STATE \textbf{end for}
\STATE \textbf{Return} node embedding $\bm{H}$, trained encoder $f$.
\end{algorithmic}
\end{algorithm}

\section{Reproducibility}
% 表x展示了我们在evaluate的时候使用的各种对比方法的源码的github 链接
Table \ref{tab:baseline} presents the GitHub links to the source codes of various contrastive methods used in our evaluation.

\bibliography{aaai25}